\documentclass{llncs}
\usepackage{amssymb,amsmath,subfigure,graphicx, pxfonts}
\usepackage{array}
\usepackage{rotating}
\usepackage{lmodern,blindtext}
\usepackage{epsfig}
\usepackage{multirow}
\usepackage{qtree}
\usepackage{stmaryrd}
\usepackage{ctable}

\begin{document}
\title{ A New Rational Algorithm for View Updating
in Relational Databases
\thanks{This paper extends work from Behrend [6] and Delhibabu [18].}}
\author{Radhakrishnan Delhibabu \inst{1} \and Andreas Behrend \inst{2}} \institute{
Informatik 5, Knowledge-Based Systems Group\\
RWTH Aachen, Germany\\
\email{delhibabu@kbsg.rwth-aachen.de}
\and Informatik 3, IDB Group\\
University of Bonn, Germany\\\email{behrend@cs.uni-bonn.de} }

\maketitle
\begin{abstract}
The dynamics of belief and knowledge is one of the major components
of any autonomous system  that should be able to incorporate new
pieces of information. In order to apply the rationality result of
belief dynamics theory to various practical problems, it should be
generalized in two respects: first it should allow a certain part of
belief to be declared as immutable; and second, the belief state
need not be deductively closed. Such a generalization of belief
dynamics, referred to as base dynamics, is presented in this paper,
along with the concept of a generalized revision algorithm for
knowledge bases (Horn or Horn logic with stratified negation). We
show that knowledge base dynamics has an interesting connection with
kernel change via hitting set and abduction. In this paper, we show
how techniques from disjunctive logic programming can be used for
efficient (deductive) database updates. The key idea is to transform
the given database together with the update request into a
disjunctive (datalog) logic program and apply disjunctive techniques
(such as minimal model reasoning) to solve the original update
problem. The approach extends and integrates standard techniques for
efficient query answering and integrity checking. The generation of
a hitting set is carried out through a hyper tableaux calculus and
magic set that is focused on the goal of minimality.

\vspace{0.5cm}

\end{abstract}
\section{Introduction}
Modeling intelligent agents' reasoning requires designing knowledge
bases for the purpose of performing symbolic reasoning. Among the
different types of knowledge representations in the domain of
artificial intelligence, logical representations stem from classical
logic. However, this is not suitable for representing or treating
items of information containing vagueness, incompleteness or
uncertainty, or knowledge base evolution that leads the agent to
change his beliefs about the world.

When a new item of information is added to a knowledge base, it may
become inconsistent. Revision means modifying the knowledge base in
order to maintain consistency, while keeping the new information and
removing (contraction) or not removing the least possible previous
information. In our case, update means revision and contraction,
that is insertion and deletion from a database perspective. Previous
work \cite{Del} and \cite{Arav1,Arav} makes connections with
revision from knowledge base dynamics.

Our knowledge base dynamics is defined in two parts: an immutable
part (formulae) and updatable part (literals) (for definition and
properties see works of Nebel \cite{Nebel} and Segerberg
\cite{Seg}). Knowledge bases have a set of integrity constraints. In
the case of finite knowledge bases, it is sometimes hard to see how
the update relations should be modified to accomplish certain
knowledge base updates.

\begin{example} Consider a database with an (immutable) rule that a staff
member is a person who is currently working in a research group
under a chair. Additional (updatable) facts are that matthias and
gerhard are group chairs, and delhibabu and aravindan are staff
members in group info1. Our first integrity constraint (IC) is that
each research group has only one chair i.e., $\forall x,y,z$ (y=z)
$\leftarrow$ group\_chair(x,y) $\wedge$ group\_chair(x,z). Second
integrity constraint is that a person can be a chair for only one
research group, i.e., $\forall x,y,z$ (y=z)$\leftarrow$
group\_chair(y,x) $\wedge$ group\_chair(z,x).

\end{example}

\begin{center}
\underline {Immutable part}: staff\_chair(X,Y)$\leftarrow$
staff\_group(X,Z),group\_chair(Z,Y). \vspace{0.5cm}

\underline{Updatable part}: group\_chair(infor1,matthias)$\leftarrow$ \\
\hspace{2.4cm}group\_chair(infor2,gerhard)$\leftarrow$ \\
\hspace{2.6cm}staff\_group(delhibabu,infor1)$\leftarrow$ \\
\hspace{2.6cm}staff\_group(aravindan,infor1)$\leftarrow$ \\
\end{center}
Suppose we want to update this database with the information,
staff\_chair({aravin\-dan},{gerhard}); From the immutable part, we
can deduce that this can be achieved by asserting
staff\_group(\underline{aravindan},Z) $\bigwedge$
group\_chair(Z,\-\underline{gerhard})

If we are restricted to definite clauses, there are three plausible
ways to do this. When dealing with the revision of a knowledge base
(both insertions and deletions), there are other ways to change a
knowledge base and it has to be performed automatically too.
Considering the information, change is precious and must be
preserved as much as possible. The \emph{principle of minimal
change} \cite{Herz,Schul} can provide a reasonable strategy. On the
other hand, practical implementations have to handle contradictory,
uncertain, or imprecise information, so several problems can arise:
how to define efficient change in the style of Carlos
Alchourr$\acute{o}$n, Peter G$\ddot{a}$rdenfors, and David Makinson
(AGM) \cite{Alch}; what result has to be chosen
\cite{Lak,Lobo,Nayak1}; and finally, according to a practical point
of view, what computational model to support for knowledge base
revision has to be provided?

The basic idea in \cite{Beh,Arav2} is to employ the model generation
property of hyper tableaux and magic set to generate models, and
read off diagnosis from them. One specific feature of this diagnosis
algorithm is the use of semantics (by transforming the system
description and the observation using an initial model of the
correctly working system) in guiding the search for a diagnosis.
This semantical guidance by program transformation turns out to be
useful for database updates as well. More specifically we use a
(least) Herbrand model of the given database to transform it along
with the update request into a disjunctive logic program in such a
way that the models of this transformed program stand for possible
updates.

We discuss two ways of transforming the given database together with
the view update (insert and delete) request into a disjunctive logic
program resulting in two variants of view update algorithms. In the
first variant, a simple and straightforward transformation is
employed. Unfortunately, not all models of the transformed program
represent a rational update using this approach. The second variant
of the algorithm uses the least Herbrand model of the given database
for the transformation. In fact what we referred to as offline
preprocessing before is exactly this computation of the least
Herbrand model. This variant is very meaningful in applications
where views are materialized for efficient query answering. The
advantage of using the least Herbrand model for the transformation
is that all models of the transformed disjunctive logic program (not
just the minimal ones) stand for a rational update.

The rest of paper is organized as follows: First we start with
preliminaries in Section 2. In Section 3, we introduce knowledge
base dynamics along with the concept of generalized revision, and
revision operator for knowledge base. Section 4 studies the
relationship between knowledge base dynamics and abduction. We
discuss an important application of knowledge base dynamics in
providing an axiomatic characterization for updating view literals
over databases. We briefly discuss hyper tableaux calculus and magic
set in Section 5. We present two variants of our rational and
efficient algorithm for view updating in Section 6. In Section 7, we
give a brief overview of related work. In Section 8 we draw
conclusions with a summary of our contribution and indicate future
directions of our investigation. All proofs can be found in the
Appendix.
\section{Background}

\subsection{Rationality of change}

Rationality of change has been studied at an abstract philosophical
level by various researchers, resulting in well known AGM Postulates
for revision \cite{Alch,Hans1}. However, it is not clear how these
rationality postulates can be applied in real world problems such as
database updates and this issue has been studied in detail by works
such as \cite{Del}. In the sequel, we briefly recall the postulates
and an algorithm for revision based on abduction from Delhibabu
\cite{Del} work.

We consider a propositional language $\mathcal{L_P}$ defined from a
finite set of propositional variables $\mathcal{P}$ and the standard
connectives. We use lower case Roman letters $a, b, x, y,...$ to
range over elementary letters and Greek letters $\varphi, \phi,
\psi, ...$ for propositional formulae. Sets of formulae are denoted
by upper case Roman letters $A,B, F,K, ....$. A literal is an atom
(positive literal), or a negation of an atom (negative literal).

Formally, a finite Horn knowledge base $KB$ (Horn \cite{Delg} or
Horn logic with stratified negation \cite{Jack}) is defined as a
finite set of formulae from language $\mathcal{L_{H}}$, and divided
into three parts: an immutable theory $KB_{I}$ is a Horn formula,
which is the fixed part of the knowledge; updatable theory $KB_{U}$
is a Horn clause; and integrity constraint $KB_{IC}$ representing a
set of clauses (Horn logic with stratified negation).

\begin{definition} [Horn Knowledge Base] \label{D1} A Horn knowledge base, KB
is a finite set of Horn formulae from language $\mathcal{L_{H}}$,
s.t $KB=KB_{I}\cup KB_{U}\cup KB_{IC}$, $KB_{I}\cap
KB_{U}=\varnothing$ and $KB_{U}\cap KB_{IC}=\varnothing$.
\end{definition}

In the AGM approach, a belief is represented by a sentence over a
suitable language $\mathcal{L_{H}}$, and a belief $KB$ is
represented by a set of sentence that are close wrt the logical
closure operator $Cn$. It is assumed that $\mathcal{L_{H}}$, is
closed under application of the boolean operators negation,
conjunction, disjunction, and implication.

\begin{definition} \label{D9} Let KB be a knowledge base with an immutable part
$KB_{I}$. Let $\alpha$ and $\beta$ be any two (Horn or Horn logic
with stratified negation) clauses from $\mathcal{L_H}$. Then,
$\alpha$ and $\beta$ are said to be \emph{KB-equivalent} iff the
following condition is satisfied: $\forall$ set of Horn clauses E
$\subseteq \mathcal{L_H}$: $KB_{I}\cup E\vdash\alpha$ iff
$KB_{I}\cup E\vdash\beta$.
\end{definition}

The revision can be trivially achieved by expansion, and the
axiomatic characterization could be straightforwardly obtained from
the corresponding characterizations of the traditional models
\cite{Fal}. The aim of our work is not to define revision from
contraction, but rather to construct and axiomatically characterize
revision operators in a direct way.

These postulates stem from three main principles: the new item of
information has to appear in the revised knowledge base, the revised
base has to be consistent and revision operation has to change the
least possible beliefs. Now we consider the revision of a Horn (Horn
logic with stratified negation) clause $\alpha$ wrt KB, written as
$KB*\alpha$. The rationality postulates for revising $\alpha$ from
KB can be formulated as follows:

\begin{definition} [Rationality postulates for knowledge base
revision] \label{10}
\begin{enumerate}
\item[]\hspace{-0.6cm}(KB*1)\hspace{0.2cm}  \emph{Closure:} $KB*\alpha$ is a knowledge base.
\item[]\hspace{-0.6cm}(KB*2)\hspace{0.2cm}  \emph{Weak Success:} if $\alpha$ is consistent with $KB_{I}\cup KB_{IC}$ then
$\alpha \subseteq KB*\alpha$.
\item[]\hspace{-0.6cm}(KB*3.1)  \emph{Inclusion:} $KB*\alpha\subseteq
Cn(KB\cup\alpha)$.
\item[]\hspace{-0.6cm}(KB*3.2)  \emph{Immutable-inclusion:} $KB_{I}\subseteq
Cn(KB*\alpha)$.
\item[]\hspace{-0.6cm}(KB*4.1)  \emph{Vacuity 1:} if $\alpha$ is
inconsistent with $KB_{I}\cup KB_{IC}$ then $KB*\alpha=KB$.
\item[]\hspace{-0.6cm}(KB*4.2)  \emph{Vacuity 2:} if $KB\cup \alpha \nvdash \perp$ then $KB*\alpha$ = $KB \cup
\alpha$.
\item[]\hspace{-0.6cm}(KB*5)\hspace{0.3cm}   \emph{Consistency:} if $\alpha$ is consistent with $KB_{I}\cup KB_{IC}$
 then $KB*\alpha$ is consistent with $KB_{I}\cup KB_{IC}$.
\item[]\hspace{-0.6cm}(KB*6)  \hspace{0.2cm} \emph{Preservation:} If $\alpha$ and $\beta$ are
KB-equivalent, then $KB*\alpha \leftrightarrow KB*\beta$.
\item[]\hspace{-0.6cm}(KB*7.1)  \emph{Strong relevance:} $KB*\alpha\vdash \alpha$ If $KB_{I}\nvdash\neg\alpha$
\item[]\hspace{-0.6cm}(KB*7.2)  \emph{Relevance:} If $\beta\in KB\backslash KB*\alpha$,
then there is a set $KB'$ such that\\ $KB*\alpha\subseteq
KB'\subseteq KB\cup\alpha$, $KB'$ is consistent $KB_{I}\cup KB_{IC}$
with $\alpha$, but $KB' \cup \{\beta\}$ is inconsistent $KB_{I}\cup
KB_{IC}$ with $\alpha$.
\item[]\hspace{-0.6cm}(KB*7.3)  \emph{Weak relevance:} If $\beta\in KB\backslash KB*\alpha$,
then there is a set $KB'$ such that $KB'\subseteq KB\cup\alpha$,
$KB'$ is consistent $KB_{I}\cup KB_{IC}$ with $\alpha$, but $KB'
\cup \{\beta\}$ is inconsistent $KB_{I}\cup KB_{IC}$ with $\alpha$.
\end{enumerate}
\end{definition}

Now we recall an algorithm for revision based on abduction presented
in \cite{Alis,Del}, Some basic definitions required for the
algorithm are presented first.

\begin{definition}[Minimal abductive explanation] \label{D20} Let KB be a Horn knowledge base and $\alpha$ an
observation to be explained. Then, for a set of abducibles
$(KB^{I})$, $\Delta$ is said to be an abductive explanation wrt
$KB^{I}$ iff $KB^{I}\cup \Delta\vdash \alpha$. $\Delta$ is said to
be \emph{minimal}\cite{Sak} wrt $KB^{I}$ iff no proper subset of
$\Delta$ is an abductive explanation for $\alpha$, i.e.,
$\nexists\Delta^{'}$ s.t. $KB^{I}\cup\Delta^{'}\vdash\alpha$.
\end{definition}

\begin{definition}[Local minimal abductive explanations] \label{D21}
Let $KB^{I'}$ be a smallest subset of $KB^{I}$, s.t $\Delta$ is a
minimal abductive explanation of $\alpha$ wrt $KB^{I'}$ (for some
$\Delta$). Then $\Delta$ is called local minimal \cite{Chris,Lu} for
$\alpha$ wrt $KB^{I}$.
\end{definition}

The general revision algorithm of \cite{Del} is reproduced here as
Algorithm 1. The basic idea behind this algorithm is to generate all
(locally minimal) explanations for the sentence to be contracted and
determine a hitting set for these explanations. Since all (locally
minimal) explanations are generated this algorithm is of exponential
space and time complexity

\begin{definition} [Hitting set]
Let S be a set of sets. Then a set HS is a \emph{hitting set} of S
iff $HS\subseteq\cup S$ and for every non-empty element R of S, $R
\cap HS$ is non empty.
\end{definition}

$$\begin{array}{cc}\hline \text{\bf Algorithm 1} & \hspace{-4cm}
\text{\rm Generalized revision algorithm}\\\hline \text{\rm Input}:&
\hspace{-0.6cm}\text{\rm A Horn knowledge base}~ KB=KB_{I}\cup KB_{U}\cup KB_{IC}\\
&\text{\rm and a Horn clause}~
\alpha~ \text{\rm to be revised.}\\
\text{\rm Output:} & \text{\rm A new Horn knowledge base
}~KB'=KB_{I}\cup
KB_{U}^*\cup KB_{IC},\\
&\text{s.t.}~ KB'\text{\rm is
a generalized revision}~ \alpha~\text{\rm to KB.}\\
\text{\rm Procedure}~KB(KB,\alpha)&\\
\text{\rm begin}&\\
~~1.&\hspace{-0.5cm}\text{\rm Let V:=}~\{c\in KB_{IC}~|~ KB_I\cup
KB_{IC}~\text{\rm inconsistent
with}~\alpha~\text{\rm wrt}~c\}\\
&P:=N:=\emptyset~\text{\rm and}~KB'=KB\\
~~2.&\text{\rm While}~(V\neq \emptyset)\\
&\text{\rm select a subset}~V'\subseteq V\\
&\text{\rm For each}~v\in~V',~\text{\rm select a literal to be}\\
&\hspace{-0.1cm}\text{\rm remove (add to N) or a literal to be added(add to P)}\\
&\text{\rm Let KB}~:=KR(KB,P,N)\\
&\hspace{-0.3cm}\text{\rm Let V:=}~\{c\in KB_{IC}~|~ KB_I~\text{\rm
inconsistent
with}~\alpha~\text{\rm wrt}~c\}\\
&\hspace{-0.7cm}\text{\rm return}\\
~~3.&\text{\rm Produce a new Horn knowledge base}~KB'\\
\text{\rm end.}&\\ \hline
\end{array}$$

\vspace{0.5cm}

$$\begin{array}{c}\hline
\text{\bf Algorithm 2} \label{A2}\\
\text{\rm Procedure}~
KR(KB,\Delta^{+},\Delta^{-})\\
\text{\rm begin}\\
1.~\text{\rm Let}~ P :=\{ e \in \Delta^{+} |~ KB_I\not\models e\}
~\text{\rm and}~ N :=\{ e \in \Delta^{-}
 |~KB_I\models e\}\\
2.~\text{\rm While}~(P\neq \emptyset)~\text{\rm or}~(N\neq \emptyset)\\
\text{\rm select a subset}~P'\subseteq P~ or ~N'\subseteq N \\
\hspace{-1.7cm}\text{\rm Construct a set}~S_1=\{X~|~X~\text{\rm is a
KB-closed
locally}\\
\text{\rm minimal abductive
wrt P explanation for}~\alpha~\text{\rm wrt}~KB_{I}\}.\\
\hspace{-1.7cm}\text{\rm Construct a set}~S_2=\{X~|~X~\text{\rm is a
KB-closed
locally}\\
\text{\rm  minimal abductive wrt N explanation for}~\alpha~\text{\rm wrt}~KB_{I}\}.\\
\text{\rm Determine a hitting set}~\sigma (S_1) \text{\rm ~and}~\sigma (S_2)\\
\hspace{-5.5cm}
\text{\rm If}~((N'=\emptyset)~and~(P'\neq\emptyset))\\
\hspace{-1cm}\text{\rm Produce}~KB'=KB_{I}\cup \{(KB_{U} \cup \sigma (S_1)\}\\
\hspace{-8.8cm}
\text{\rm else}\\
\text{\rm Produce}~KB'=KB_{I}\cup \{(KB_{U}\backslash
\sigma(S_2) \cup \sigma (S_1)\}\\
\hspace{-8.5cm}
\text{\rm end if}\\
\hspace{-5.5cm}
\text{\rm If}~((N'\neq\emptyset)~\text{\rm and}~(P'=\emptyset))\\
\hspace{-1.2cm}\text{\rm Produce}~KB'=KB_{I}\cup \{(KB_{U}\backslash
\sigma(S_2)\}\\
\hspace{-8.8cm}
\text{\rm else}\\
\text{\rm Produce}~KB'=KB_{I}\cup \{(KB_{U}\backslash
\sigma(S_2) \cup \sigma (S_1)\}\\
\hspace{-8.5cm}
\text{\rm end if}\\
 \text{\rm Let}~ P :=\{ e \in \Delta^{+} |~ KB_I\not\models e\}
~\text{\rm and}~ N :=\{ e \in \Delta^{-}
 |~KB_I\models e\}\\
3.~\text{\rm return}~ KB'\\
\text{\rm end.}\\ \hline
\end{array}$$

\begin{theorem} \label{T8} Let KB be a knowledge base and $\alpha$
is (Horn or Horn logic with stratified negation) formula.
\begin{enumerate}
  \item If Algorithm 1 produced KB' as a result of revising $\alpha$
  from KB, then KB' satisfies all the rationality postulates (KB*1) to
(KB*6) and (KB*7.3).
  \item Suppose $KB''$ satisfies all these rationality postulates
  for revising $\alpha$ from KB, then $KB''$ can be produced by Algorithm 1.
\end{enumerate}
\end{theorem}


\section{Deductive database} A \it Deductive database \rm $DDB$ consists of three parts:
an \it intensional database \rm $IDB$ ($KB_I$), a set of definite
program clauses, \it extensional database \rm $EDB$ ($KB_U$), a set
of ground facts; and \it integrity constraints\rm ~$IC$. The
intuitive meaning of $DDB$ is provided by the \it Least Herbrand
model semantics \rm and all the inferences are carried out through
\it SLD-derivation. All the predicates that are defined in $IDB$ are
referred to as \emph{view predicates}  and those defined in $EDB$
are referred to as \emph{base predicates}. \rm Extending this
notion, an atom with a view predicate is said to be a \it view
atom,\rm and similarly an atom with base predicate is a \it base
atom. \rm Further we assume that $IDB$ does not contain any unit
clauses and no predicate defined in a given $DDB$ is both view and
base.

Two kinds of view updates can be carried out on a $DDB$: An atom,
that does not currently follow from $DDB$, can be \it inserted, \rm
or an atom, that currently follows from $DDB$ can be \it deleted.
 \rm When an atom $A$ is to be updated, the view update problem is to
insert or delete only some relevant $EDB$ facts, so that the
modified $EDB$ together with $IDB$ will satisfy the updating of $A$
to $DDB$.

Note that a $DDB$ can be considered as a knowledge base to be
revised. The $IDB$ is the immutable part of the knowledge base,
while the $EDB$ forms the updatable part. In general, it is assumed
that the language underlying a $DDB$ is fixed and the semantics of
$DDB$ is the least Herbrand model over this fixed language. We
assume that there are no function symbols implying that the Herbrand
Base is finite. Therefore, the $IDB$ is practically a shorthand of
its ground instantiation\footnotemark \footnotetext{a ground
instantiation of a definite program $P$ is the set of clauses
obtained by substituting terms in the Herbrand Universe for
variables in $P$ in all possible ways}  written as $IDB_G$. In the
sequel, technically we mean $IDB_G$ when we refer simply to $IDB$.
Thus, a $DDB$ represents a knowledge base where the immutable part
is given by $IDB_G$ and updatable part is the $EDB$. Hence, the
rationality postulates (KB*1)-(KB*6) and (KB*7.3) provide an
axiomatic characterization for update (insert and delete) a view
atom $A$ from a definite database $DDB$.

Logic provides a conceptual level for understanding the meaning of
relational databases. Hence, the rationality postulates
(KB*1)-(KB*6) and (KB*7.3) can provide an axiomatic characterization
for view updates in relational databases too. A relational database
together with its view definitions can be represented by a deductive
database ($EDB$ representing tuples in the database and $IDB$
representing the view definitions), and so the same algorithm can be
used to delete view extensions from relational deductive databases.

\subsection{Disjunctive Deductive Databases} A disjunctive Datalog
rule is a function-free clause of the form $H_1 \lor\ldots\lor H_m
\leftarrow L_1 \land\ldots \land L_n$ with $m, n\geq 1$ where the
rule's head $H_1 \lor\ldots\lor H_m$ is a disjunction of positive
atoms, and the rule's body $L_1 \land\ldots\land L_n$ consists of
literals, i.e., positive or negative atoms, if only positive atoms
then (definite deductive) database. If $H\equiv H_1 \lor\ldots\lor
H_m$ is the head of a given rule $IDB$, we use $pred(IDB)$ to refer
to the set of predicate symbols of $H$, i.e.,
$pred(IDB)=\{pred(H_1),\ldots, pred(H_m)\}$. For a set of rules
$IDB$, $pred(IDB)$ is defined again as $\bigcup_{r\in IDB} pred(r)$.
A \emph{disjunctive fact} $f \equiv f_1\lor \ldots \lor f_k$ is a
disjunction of ground atoms $f_i$ with $i\geq 1$. $f$ is called
\emph{definite} if $i = 1$. In the following, we identify a
disjunctive fact with a set of atoms such that the occurrence of a
ground atom $A$ within a fact $f$ can also be written as $A\in f$.
The set difference operator can then be used to exclude certain
atoms from a disjunction while the empty set is interpreted as the
boolean constant false.

A disjunctive deductive database $DDDB$ is a pair $\langle
IDB,EDB,IC \rangle$ where $EDB$ is a finite set of disjunctive facts
and $IDB$ a finite set of disjunctive rules such that $pred(EDB)\cap
pred(IDB) = \emptyset$. Again, stratifiable (definite) deductive
rules are considered only, that is, recursion through negative
predicate occurrences is not permitted. In addition to the usual
stratification concept for definite rules it is required that all
predicates within a rule's head are assigned to the same stratum.

An update request U = B, where B is a set of base facts, is not true
in KB. Then, we need to find a transaction $T=T_{ins} \cup T_{del}$,
where $T_{ins} (\Delta_i)$ (resp. $T_{del}(\Delta_j)$) is the set of
facts, such that U is true in $DDB'=((EDB - T_{del} \cup T_{ins})
\cup IDB \cup IC)$. Since we consider stratifiable (definite)
deductive databases, SLD-trees can be used to compute the required
abductive explanations. The idea is to get all EDB facts used in a
SLD-derivation of $A$ wrt DDB, and construct that as an abductive
explanation for $A$ wrt $IDB_G$.

All solutions translate a view update request into a
\textbf{transaction combining insertions and deletions of base
relations} for satisfying the request \cite{Mota}. Furthermore, a
stratifiable (definite) deductive database can be considered as a
knowledge base, and thus the rationality postulates and insertion
algorithm from the previous section can be applied for solving view
update requests in deductive databases.

\newpage

\begin{example} \label{E12}
Consider a definite deductive database DDB as follows:

$$\begin{array}{cccccc} IDB:&p\leftarrow
a\wedge e&\hspace{0.5cm}EDB:&a\leftarrow&\hspace{1.2cm}IC:&\leftarrow b\\
&q\leftarrow a\wedge f&&e\leftarrow&&\\
&p\leftarrow b\wedge f&&f\leftarrow&&\\
&q\leftarrow b\wedge e&&&&\\
&\hspace{-0.6cm}p\leftarrow q&&&&\\
&\hspace{-0.6cm}q\leftarrow a&&&&\end{array}$$

Suppose We want to insert $p$. First, we need to check consistency
with IC and afterwards, we have to find $\Delta_{i}$ and
$\Delta_{j}$ via tree deduction.

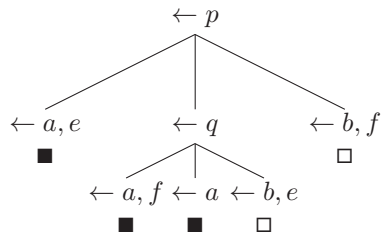
\Tree[ {$\leftarrow a,e$\\$\blacksquare$} [.$\leftarrow q$
{$\leftarrow a,f$\\$\blacksquare$} {$\leftarrow a$\\$\blacksquare$}
{$\leftarrow b,e$\\$\Box$} ].$\leftarrow q$ {$\leftarrow
b,f$\\$\Box$} ].$\leftarrow p$

\end{example}

It is easy to conclude which branches are consistent wrt IC
(indicated in the depicted tree by the symbol $\blacksquare$). For
the next step, we need to find minimal accommodate (positive
literal) and denial literal (negative literal) with wrt to $p$. The
subgoals of the tree are $\leftarrow a,e$ and $\leftarrow a,f$,
which are minimal tree deductions of only facts. Clearly,
$\Delta_{i} =\{a,e,f\}$ and $\Delta_{j} =\{b\}$ with respect to IC,
are the only locally minimal abductive explanations for $p$ wrt
$IDB_G$, but they are not locally minimal explanations.

An algorithm for view update, based on the general revision
algorithm (cf. Algorithm 1) in Section 2.1 and abductive
explanation. There, given a view atom to be updated, set of all
explanations for that atom has to be generated through a complete
$SLD$-tree and a hitting set of these explanations is then update
from the $EDB$. It was shown that this algorithm is rational. In
this paper, we present a radically different approach that runs on
polynomial space. The generation of hitting set is carried out
through a hyper tableaux calculus (bottom-up) for deletion process
and magic set(top-down) for insertion are focussed on the goal.

An algorithm for view updating can be developed based on the general
revision algorithm and the generation of abductive explanations as
proposed by Algorithm 1 and 2. For processing a given view update
request, a set of all explanations for that atom has to be generated
through a complete $SLD$-tree. The resulting hitting set of these
explanations is then a base update of the $EDB$ satsifying the view
update request. In \cite{Arav2}, it has been shown that this
algorithm is rational. In this paper, we present a different
approach which is also rational but even runs on polynomial space.
The generation of a hitting set is carried out through a hyper
tableaux calculus (bottom-up) for implementing the deletion process
as well as through the magic sets approach (top-down) for performing
insertions focussed on the particular goal given.


\subsection{View update method}

View updating \cite{Beh} aims at determining one or more base
relation updates such that all given update requests with respect to
derived relations are satisfied after the base updates have been
successfully applied.

\begin{definition}[View update] Let $DDB = \langle IDB,EDB,IC\rangle$ be a
stratifiable (definite) deductive database $DDB(D)$. A VU request
$\nu_{D}$ is a pair $\langle \nu^+_{D},\nu^-_{D}\rangle$ where
$\nu^+_{D}$ and $\nu^-_{D}$ are sets of ground atoms representing
the facts to be inserted into $D$ or deleted from $D$, resp., such
that $pred(\nu^+_{D}\cup \nu^-_{D}) \subseteq pred(IDB)$,
$\nu^+_{D}\cap \nu^-_{D} = \emptyset$, $\nu^+_{D}\cap
PM_{D}=\emptyset$ and $\nu^-_{D}\subseteq PM_{D}$.\rm
\end{definition}

Note that we consider again true view updates only, i.e., ground
atoms which are presently not derivable for atoms to be inserted, or
are derivable for atoms to be deleted, respectively. A method for
view updating determines sets of alternative updates satisfying a
given request. A set of updates leaving the given database
consistent after its execution is called \emph{VU realization}.

\begin{definition} [Induced update] Let $DDB = \langle IDB,EDB,
IC\rangle$ be a stratifiable (definite) deductive database and
$DDB=\nu_{D}$ a VU request. A VU realization is a base update
$u_{D}$ which leads to an induced update $u_{D\rightarrow D'}$ from
$D$ to $D'$ such that $\nu^+_{D}\subseteq PM_{D'}$ and
$\nu^-_{D}\cap PM_{D'}=\emptyset$.\rm
\end{definition}

There may be infinitely many realizations and even realizations of
infinite size which satisfy a given VU request. A breadth-first
search (BFS) is employed for determining a set of minimal
realizations $\tau_{D}= \{u^1_{D},\ldots, u^i_{D}\}$. Any $u^i_{D}$
is minimal in the sense that none of its updates can be removed
without losing the property of being a realization for $\nu_{D}$.

\subsubsection{Top-down computation:} Given a VU request $\nu_{DDB}$, view
updating methods usually determine further VU requests in order to
find relevant base updates. Similar to delta relations for UP we
will use the notion VU relation to access individual view updates
with respect to the relations of our system. For each relation $p\in
pred(IDB\cup EDB)$ we use the VU relation $\nabla^+_ p(\vec{x})$ for
tuples to be inserted into $DDB$ and $\nabla^-_ p(\vec{x})$ for
tuples to be deleted from $DDB$. The initial set of delta facts
resulting from a given VU request is again represented by so-called
\emph{VU seeds}.

\begin{definition}[View update seeds]  Let $DDB(D)$ be a stratifiable (definite) deductive database and
$\nu_{DDB}=\langle \nu^+_{D},\nu^-_{D}\rangle$ a VU request. The set
of VU seeds $vu\_seeds(\nu_D)$ with respect to $\nu_D$ is defined as
follows:
$$vu\_seeds(\nu_D) := \left\{\nabla^{\pi}_p (c_1,\ldots , c_n)~|~p(c_1,\ldots, c_n)\in \nu^{\pi}_D~and~\pi\in\{+,
-\}\right\} .$$ \rm
\end{definition}

\begin{definition}[View update rules]  Let $IDB$ be a normalized
stratifiable (definite) deductive rule set. The set of VU rules for
true view updates is denoted $IDB^{\nabla}$ and is defined as the
smallest set satisfying the following conditions:
\begin{enumerate}
\item[1.] For each rule of the
form $p(\vec{x})\leftarrow q(\vec{y})\land r(\vec{z})\in IDB$ with
$vars(p(\vec{x})) = (vars(q(\vec{y}))\cup vars(r(\vec{z})))$ the
following three VU rules are in $IDB^{\nabla}$:
$$\begin{array}{ccc} \nabla^+_p(\vec{x})\land \neg
q(\vec{y})\rightarrow \nabla^+_ q (\vec{y})&~~& \nabla^-_ p
(\vec{x})
\rightarrow \nabla^-_ q (\vec{y})\lor \nabla^-_ r (\vec{z})\\
\nabla^+_ p (\vec{x}) \land \neg r(\vec{z}) \rightarrow\nabla^+_ r
(\vec{z})&~~&\end{array}$$
\item[2.] For each rule of the form $p(\vec{x})\leftarrow
q(\vec{x})\land \neg r(\vec{x})\in IDB$ the following three VU rules
are in $IDB^{\nabla}$: $$\begin{array}{ccc} \nabla^+_ p
(\vec{x})\land\neg q(\vec{x}) \rightarrow\nabla^+_ q (\vec{x})&~~&
\nabla^- _p (\vec{x}) \rightarrow \nabla^-_ q (\vec{x}) \lor
\nabla^+_ r (\vec{x})\\ \nabla^+_ p (\vec{x})\land r(\vec{x})
\rightarrow \nabla^-_ r (\vec{x})&~~&\end{array}$$
\item[3.] For each two rules of the form $p(\vec{x})\leftarrow q(\vec{x})$ and
$p(\vec{x})\leftarrow r(\vec{x})$ the following three VU rules are
in $IDB^{\nabla}$: $$\begin{array}{ccc} \nabla^-_ p (\vec{x})\land
q(\vec{x}) \rightarrow\nabla^-_ q (\vec{x})&~~& \nabla^+_ p
(\vec{x}) \rightarrow \nabla^+_ q (\vec{x}) \lor \nabla^+_ r
(\vec{x})\\ \nabla^-_ p (\vec{x}) \land r(\vec{x})
\rightarrow\nabla^-_ r (\vec{x})&~~&\end{array}$$
\item[4.]\begin{enumerate} \item[a)] For each relation $p$ defined
by a single rule $p(\vec{x}) \leftarrow  q(\vec{y})\in IDB$ with
$vars(p(\vec{x})) = vars(q(\vec{y}))$ the following two VU rules are
in $IDB^{\nabla}$:
$$\begin{array}{ccc}
 \nabla^+_ p (\vec{x}) \rightarrow \nabla^+_ q (\vec{y})&~~& \nabla^-_ p (\vec{x}) \rightarrow \nabla^-_ q
(\vec{y})\end{array}$$
\item[b)] For each relation $p$ defined by a
single rule $p \leftarrow \neg q \in IDB$ the following two VU rules
are in $IDB^{\nabla}$: $$\begin{array}{ccc}\nabla^+_ p \rightarrow
\nabla^-_ q &~~& \nabla^-_ p \rightarrow \nabla^+_ q
\end{array}$$
\end{enumerate}
\item[5.] Assume without loss of
generality that each projection rule in $IDB$ is of the form
$p(\vec{x})\leftarrow q(\vec{x}, Y )\in IDB$ with $Y \notin
vars(p(\vec{x}))$. Then the following two VU rules
\begin{eqnarray*} &&\nabla^- p_(\vec{x})\land q(\vec{x}, Y )\rightarrow \nabla^-_ q (\vec{x}, Y
)\\
&&\nabla^+_ p (\vec{x})\rightarrow\nabla^+_ q (\vec{x},
c_1)\lor\ldots\lor \nabla^+_ q (\vec{x}, c_n) \lor \nabla^+_ q
(\vec{x}, c^{new})\end{eqnarray*} are in $IDB^{\nabla}$ where all
$c_i$ are constants from the Herbrand universe $\mathcal{U}_{DDB}$
of $DDB$ and $c^{new}$ is a new constant, i.e., $c^{new} \notin
\mathcal{U}_{DDB}$.
\end{enumerate}
\end{definition}

\begin{theorem} Let $DDB = \langle IDB,EDB,IC\rangle$ be a
stratifiable (definite)deductive database(D), $\nu_{D}$ a view
update request and $\tau_{D} = \{u^1_{D},\ldots, u^n_{D}\}$ the
corresponding set of minimal realizations. Let ${D}^{\nabla} =
\langle EDB \cup vu\_seeds(\nu_{D}), IDB \cup IDB^{\nabla}\rangle$
be the transformed deductive database of ${D}$. Then the VU
relations in ${PM_D^{\nabla}}$ with respect to base relations of
${D}$ correctly represent all direct consequences of $\nu_{D}$. That
is, for each realization $u^i_{D} = \langle u^{i^+}_{D} ,
u^{i^{-}}_{D}\rangle \in \tau_{D}$ the following condition holds:
\begin{equation*} \exists p(\vec{t})\in u^{i^+}_{D} : \nabla^+_ p
(\vec{t} )\in {MS_D^{\nabla}} \lor \exists p(\vec{t} )\in
u^{i^{-}}_{D} : \nabla^-_p (\vec{t} ) \in {MS_D^{\nabla}}.
\end{equation*}
\end{theorem}

\subsubsection{Bottom-up computation:}
In \cite{Bau,Arav2} a variant of clausal normal form tableaux called
"hyper tableaux" is introduced. Since the hyper tableaux calculus
constitutes the basis for our view update algorithm, \it Clauses,
\rm i.e., multisets of literals, are usually written as the
disjunction $A_1\lor A_2\lor\cdots\lor A_m\lor~\text{not}~B_1\lor
~\text{not}~B_2\cdots\lor~\text{not}~B_n$ ($M\geq 0,n\geq 0$). The
literals $A_1,A_2,\ldots A_m$ (resp. $B_1,B_2,\ldots, B_n$) are
called the \it head (\rm resp. \it body) \rm of a clause. With
$\overline{L}$ we denote the complement of a literal $L$. Two
literals $L$ and $K$ are complementary if $\overline{L}=K$.

From now on $D$ always denotes a finite ground clause set, also
called \it database, \rm and $\Sigma$ denotes its signature, i.e.,
the set of all predicate symbols occurring in it. We consider finite
ordered trees $T$ where the nodes, except the root node, are labeled
with literals. In the following we will represent a branch $b$ in
$T$ by the sequence $b=L_1,L_2,\ldots, L_n$ ($n\geq 0$) of its
literal labels, where $L_1$ labels an immediate successor of the
root node, and $L_n$ labels the leaf of $b$. The branch $b$ is
called \it regular \rm iff $L_i\neq L_j$ for $1\leq i,j\leq n$ and
$i\neq j$, otherwise it is called \it irregular. \rm The tree $T$ is
\it regular\rm~iff every of its branches is regular, otherwise it is
\it irregular. \rm The set of \it branch literals \rm of $b$ is
$lit(b)= \{L_1,L_2,\ldots,L_n\}$. For brevity, we will write
expressions like $A \in b$ instead of $A\in lit(b)$.  In order to
memorize the fact that a branch contains a contradiction, we allow
to label a branch as either \it open \rm or \it closed. \rm A
tableau is closed if each of its branches is closed, otherwise it is
open.

\begin{definition} [Hyper Tableau]
A literal set is called \it inconsistent \rm iff it contains a pair
of complementary literals, otherwise it is called \it consistent.
Hyper tableaux \rm for $D$ are inductively defined as follows: \bf

Initialization step: \rm  The empty tree, consisting of the root
node only, is a hyper tableau for $D$. Its single branch is marked
as "open". \bf

Hyper extension step: \rm  If (1) $T$ is an open hyper tableau for
$D$ with open branch $b$,  and (2) $C=A_1\lor A_2\lor\cdots\lor
A_m\leftarrow B_1\land B_2\cdots\land B_n$ is  a clause from $D$
($n\geq 0,m\geq 0$), called \it extending clause \rm in this
context, and (3) $\{B_1,B_2,\ldots, B_n\}\subseteq b$ (equivalently,
we say that $C$ is \it applicable to $b$)\rm  then the tree $T$ is a
hyper tableau for $D$, where $T$ is obtained from $T$ by extension
of $b$ by $C$:
 replace $b$ in $T$ by the \it new branches \rm \begin{equation*}
 (b,A_1),(b,A_2),\ldots,(b,A_m),(b,\neg B_1),(b,\neg B_2),\ldots,
 (b,\neg B_n)
 \end{equation*} and then mark every inconsistent new branch as "closed", and the
other new branches as "open".
\end{definition}

The applicability condition of an extension expresses that all body
literals have to be satisfied by the branch to be extended. From now
on, we consider only regular hyper tableaux. This restriction
guarantees that for finite clause sets no branch can be extended
infinitely often. Hence, in particular, no open finished branch can
be extended any further. This fact will be made use of below
occasionally. Notice as an immediate consequence of the above
definition that open branches never contain negative literals.

\section{View update algorithm}

The key idea of the algorithm presented in this paper is to
transform the given database along with the view update request into
a disjunctive logic program and apply known disjunctive techniques
to solve the original view update problem. The intuition behind the
transformation is to obtain a disjunctive logic program in such a
way that each (minimal) model of this transformed program represent
a way to update the given view atom. We present two variants of our
algorithm. The one that is discussed in this section employs a
trivial transformation procedure but has to look for minimal models;
and another performs a costly transformation, but dispenses with the
requirement of computing the minimal models.

\subsection{Minimality test}
We start presenting an algorithm for stratifiable (definite)
deductive databases by first defining precisely how the given
database is transformed into a disjunctive logic program for the
view deletion process \cite{Arav2} (successful branch - see in
\cite{Del} via Hyper Tableau).

\begin{definition} [$IDB$ Transformation] Given an $IDB$ and a set of ground atoms $S$, the
transformation of $IDB$ wrt $S$ is obtained by translating each
clause $C\in IDB$ as follows: Every atom $A$ in the body (resp.
head) of $C$ that is also in $S$ is moved to the head (resp. body)
as $\neg A$.
\end{definition}

\begin{note} If $IDB$ is a stratifiable deductive database then the transformation
  introduced above is not necessary.
\end{note}

\begin{definition} [$IDB^*$ Transformation] Let $IDB\cup EDB$ be a given database.
Let $S_0=EDB\cup \{A~|~A~\text{ is a ground \it IDB \rm atom}\}$.
Then, $IDB^*$ is defined as the transformation of $IDB$ wrt $S_0$.
\end{definition}

\begin{note} Note that $IDB^*$ is in general a disjunctive logic program.
The negative literals $(\neg A)$ appearing in the clauses are
intuitively interpreted as deletion of the corresponding atom ($A$)
from the database. Technically, a literal $\neg A$ is to be read as
a \it positive \rm atom, by taking the $\neg$-sign as part of the
predicate symbol. To be more precise, we treat $\neg A$ as an atom
wrt $IDB^*$, but as a negative literal wrt $IDB$.

Note that there are no facts in $IDB^*$. So when we add a delete
request such as $\neg A$ to this, the added request is the only fact
and any bottom-up reasoning strategy is fully focused on the goal
(here the delete request)
\end{note}

\begin{definition}[Update Tableaux Hitting Set] An update tableau for a database
$IDB\cup EDB$ and delete request $\neg A$ is a hyper tableau $T$ for
$IDB^*\cup\{\neg A\leftarrow\}$ such that every open branch is
finished. For every open finished branch $b$ in $T$ we define the
\it hitting set (of b in $T$) \rm as $HS(b)=\{A \in EDB | \neg A \in
b\}$.
\end{definition}

\begin{definition}[Minimality test] Let $T$ be an update tableau for $IDB\cup
EDB$ and delete request $\neg A$. We say that open finished branch
$b$ in $T$ satisfies \it the strong minimality test \rm iff $\forall
s\in HS(b):IDB\cup EDB\backslash HS(b)\cup\{s\}\vdash A$.
\end{definition}

\begin{definition}[Update Tableau satisfying strong minimality] An update tableau
for given $IDB\cup  EDB$ and delete request $\neg A$ is transformed
into an update tableau satisfying strong minimality by marking every
open finished branch as closed which does not satisfy strong
minimality.
\end{definition}

The next step is to consider the view insertion process \cite{Beh}
(unsuccessful branch - see \cite{Del}).

\begin{definition} [$IDB^**$ Transformation] Let $IDB\cup EDB$ be a given database.
Let $S_1=EDB\cup \{A~|~A~\text{ is a ground \it IDB \rm atom}\}$.
Then, $IDB^**$ is defined as the transformation of $IDB$ wrt $S_1$.
\end{definition}

\begin{note} Note that $IDB$ is in general a (stratifiable) disjunctive logic program.
The positive literals $(A)$ appearing in the clauses are intuitively
interpreted as an insertion of the corresponding atom ($A$) from the
database. \end{note}

\begin{definition}[Update magic Hitting Set] An update magic set rule for a database
$IDB\cup EDB$ and insertion request $A$ is a magic set rule $M$ for
$IDB^*\cup\{A\leftarrow\}$ such that every close branch is finished.
For every close finished branch $b$ in $M$ we define the \it magic
set rule (of b in $M$) \rm as $HS(b)=\{A \in EDB | A \in b\}$.
\end{definition}

\begin{definition}[Minimality test] Let $M$ be an update magic set rule for $IDB\cup
EDB$ and insert request $A$. We say that close finished branch $b$
in $M$ satisfies \it the strong minimality test \rm iff $\forall
s\in HS(b):IDB\cup EDB\backslash HS(b)\cup\{s\}\vdash \neg A$.
\end{definition}

\begin{definition}[Update magic set rule satisfying strong minimality] An update
magic set rule for given $IDB\cup  EDB$ and insert request $A$ is
transformed into an update magic set rule satisfying strong
minimality by marking every close finished branch as open which does
not satisfy strong minimality.
\end{definition}

$$\begin{array}{cc}\hline
\text{\bf Algorithm 3} &  \text{\rm View updating Algorithm based on
minimality test}\\\hline \text{\rm Input}:& \text{\rm A definite
deductive database}~ DDB=IDB\cup EDB\cup IC\\
\text{\rm Output:}&\text{\rm A new database}~IDB\cup EDB'\cup IC\\
\text{\rm begin}&\\
~~1.&\text{\rm Let}~ V :=\{ c\in IC~|~IDB\cup IC~\text{\rm
inconsistent
with}~\mathcal{A}~\text{\rm wrt}~c~\}\\
&\text{\rm While}~(V\neq \emptyset)\\
~~2.&\hspace{-0.7cm}\text{\rm For every successful branch $i$:construct}~\Delta_{i}=\{D~|~D\in EDB \}\\
&\text{\rm and D is used as an input clause in branch $i$}.\\
&\text{\rm Construct a branch i of an update tableau satisfying
minimality}\\
&\text{\rm for}~ IDB\cup EDB~\text{\rm and delete request}~\neg A.\\
&\text{\rm  Produce}~IDB\cup EDB \backslash HS(i)~\text{\rm as a result}\\
~~3.&\text{\rm For every unsuccessful branch $j$:construct}~\Delta_{j}=\{D~|~D\in EDB \}\\
&\text{\rm and D is used as an input clause in branch $j$}.\\
&\text{\rm Construct a branch j of an update magic set rule
satisfying minimality}\\
&\text{\rm for}~ IDB\cup EDB~\text{\rm and insert request}~A.\\
&\text{\rm  Produce}~IDB\cup EDB \backslash HS(j)~\text{\rm as a
result}\\
&\text{\rm Let}~ V :=\{ c\in IC~|~IDB\cup IC~\text{\rm inconsistent
with}~\mathcal{A}~\text{\rm wrt}~c~\}\\
&\hspace{-0.7cm}\text{\rm return}\\
~~5.&\text{\rm Produce}~DDB~\text{\rm as the result.}\\
\text{\rm end.}&\\\hline
\end{array}$$

$$\begin{array}{cc}\hline
\text{\bf Algorithm 4} &  \text{\rm View updating Algorithm based on
minimality test}\\\hline \text{\rm Input}:& \text{\rm A stratifiable
deductive database}~ DDB=IDB\cup EDB\cup IC\\
\text{\rm Output:}&\text{\rm A new database}~IDB\cup EDB'\cup IC\\
\text{\rm begin}&\\
~~1.&\text{\rm Let}~ V :=\{ c\in IC~|~IDB\cup IC~\text{\rm
inconsistent
with}~\mathcal{A}~\text{\rm wrt}~c~\}\\
&\text{\rm While}~(V\neq \emptyset)\\
~~2.&\hspace{-0.7cm}\text{\rm For every successful branch $i$:construct}~\Delta_{i}=\{D~|~D\in EDB \}\\
&\text{\rm and D is used as an input clause in branch $i$}.\\
&\text{\rm Construct a branch i of an update tableau satisfying
minimality}\\
&\text{\rm for}~ IDB\cup EDB~\text{\rm and delete request}~ A.\\
&\text{\rm  Produce}~IDB\cup EDB \backslash HS(i)~\text{\rm as a result}\\
~~3.&\text{\rm For every unsuccessful branch $j$:construct}~\Delta_{j}=\{D~|~D\in EDB \}\\
&\text{\rm and D is used as an input clause in branch $j$}.\\
&\text{\rm Construct a branch j of an update magic set rule
satisfying minimality}\\
&\text{\rm for}~ IDB\cup EDB~\text{\rm and insert request}~A.\\
&\text{\rm  Produce}~IDB\cup EDB \backslash HS(j)~\text{\rm as a
result}\\
&\text{\rm Let}~ V :=\{ c\in IC~|~IDB\cup IC~\text{\rm inconsistent
with}~\mathcal{A}~\text{\rm wrt}~c~\}\\
&\hspace{-0.7cm}\text{\rm return}\\
~~5.&\text{\rm Produce}~DDB~\text{\rm as the result.}\\
\text{\rm end.}&\\\hline
\end{array}$$

\newpage
\begin{lemma}
The strong minimality test and the groundedness test are equivalent.
\end{lemma}

This means that every minimal model (minimal wrt the base atoms) of
$IDB^* \cup \{\neg A\}$ provides a minimal hitting set for deleting
the ground view atom $A$. Similarly, $IDB^* \cup \{A\}$ provides a
minimal hitting set for inserting the ground view atom $A$. Now we
are in a position to formally present our algorithm. Given a
database and a view atom to be updated, we first transform the
database into a definite disjunctive logic program and use hyper
tableaux calculus to generate models of this transformed program for
deletion of an atom.  Second, magic sets transformed rules are used
is used to generate models of this transformed program for
determining an induced insertion of an atom. Models that do not
represent rational update are filtered out using the strong
minimality test. This is formalized in Algorithm 3. The procedure
for stratifiable deductive databases is presented in Algorithm 4.

To show the rationality of this approach, we study how this is
related to the previous approach presented in the last section,
i.e., generating explanations and computing hitting sets of these
explanations. To better understand the relationship it is imperative
to study where the explanations are in the hyper tableau approach
and magic set rules. We first define the notion of an $EDB$ -cut and
then view update seeds.

\begin{definition}[$EDB$-Cut] Let $T$ be update tableau with open branches
$b_1,b_2,\ldots,b_n$. A set $S=\{A_1,A_2,\ldots,A_n\}\subseteq EDB$
is said to be $EDB$-cut of $T$ iff  $\neg A_i\in b_i$ ($A_i\in
b_i$), for $1\leq i\leq n$.
\end{definition}

\begin{definition}[$EDB$ seeds]
Let $M$ be an update seeds with close branches $b_1,b_2,\ldots,b_n$.
A set $S=\{A_1,A_2,\ldots,A_n\}\subseteq EDB$ is said to be a
$EDB$-seeds of $M$ iff EDB seeds $vu\_seeds(\nu_D)$ with respect to
$\nu_D$ is defined as follows:
$$vu\_seeds(\nu_D) := \left\{\nabla^{\pi}_p (c_1,\ldots , c_n) | p(c_1,\ldots, c_n)\in \nu^{\pi}_D~and~\pi\in\{+,
-\}\right\} .$$ \rm
\end{definition}

\begin{lemma} Let $T$ be an update tableau for $IDB\cup EDB$ and
update request $A$. Similarly, for $M$ be an update magic set rule.
Let $S$ be the set of all $EDB$-closed minimal abductive
explanations for $A$ wrt. $IDB$.  Let $S'$ be the set of all
$EDB$-cuts of $T$ and $EDB$-seeds of $M$ . Then the following hold
\begin{enumerate}
\item[$\bullet$] $S\subseteq S'$.\\
\item[$\bullet$] $\forall \Delta'\in S':\exists \Delta\in S s.t. \Delta\subseteq \Delta'$.
\end{enumerate}
\end{lemma}

The above lemma precisely characterizes what explanations are
generated by an update tableau. It is obvious then that a branch
cuts through all the explanations and constitutes a hitting set for
all the generated explanations. This is formalized below.

\begin{lemma} Let $S$ and $S'$ be sets of sets s.t. $S\subseteq S'$ and
every member  of $S'\backslash S$ contains an element of S. Then, a
set $H$ is a minimal hitting set for $S$ iff it is a minimal hitting
set for $S'$.
\end{lemma}

\begin{lemma}Let $T$ be an update tableau for $IDB\cup EDB$ and
update request $A$ that satisfies the strong minimality test.
Similarly, for $M$ be an update magic set rule. Then, for every open
(close) finished branch $b$ in $T$, $HS(b)$ ($M$, $HS(b)$) is a
minimal hitting set of all the abductive explanations of $A$.
\end{lemma}

So, Algorithms 3 and 4 generate a minimal hitting set (in polynomial
space) of all $EDB$-closed locally minimal abductive explanations of
the view atom to be deleted. From the belief dynamics results
recalled in section 3, it immediately follows that Algorithms 5 and
6 are rational, and satisfy the strong relevance postulate (KB-7.1).

\begin{theorem} Algorithms 3 and 4 are rational, in the sense that
they satisfy all the rationality postulates (KB*1)-(KB*6) and the
strong relevance postulate (KB*7.1). Further, any update that
satisfies these postulates can be computed by these algorithms.
\end{theorem}

\subsection{Materialized view}
In many cases, the view to be updates is materialized, i.e., the
least Herbrand Model is computed and kept, for efficient query
answering. In such a situation, rational hitting sets can be
computed without performing any minimality test. The idea is to
transform the given $IDB$ wrt the materialized view.

\begin{definition} [$IDB^+$ Transformation] Let $IDB\cup  EDB$ be a given
database. Let $S$ be the Least Herbrand Model of this database.
Then, $IDB^+$ is defined as the transformation of $IDB$ wrt $S$.
\end{definition}

\begin{note}
If $IDB$ is a stratifiable deductive database then the
transformation introduced above is not necessary.
\end{note}

\begin{definition}[Update Tableau based on Materialized view] An update tableau
based on materialized view for a database $IDB\cup EDB$ and delete
request $\neg A$ is a hyper tableau $T$ for $IDB^+\cup\{\neg
A\leftarrow \}$ such that  every open branch is finished.
\end{definition}

\begin{definition} [$IDB^-$ Transformation] Let $IDB\cup  EDB$ be a given
database. Let $S_1$ be the Least Herbrand Model of this database.
Then, $IDB^-$ is defined as the transformation of $IDB$ wrt $S_1$.
\end{definition}

\begin{definition}[Update magic set rule based on Materialized view] An update
magic set rule based on materialized view for a database $IDB\cup
EDB$ and insert request $A$ is a magic set $M$ for
$IDB^+\cup\{A\leftarrow \}$ such that  every close branch is
finished.
\end{definition}

Now the claim is that every model of $IDB^+\cup\{\neg A\leftarrow
\}$ ($A\leftarrow$) constitutes a rational hitting set for the
deletion and insertion of the ground view atom $A$. So, the
algorithm works as follows:

$$\begin{array}{cc}\hline
\text{\bf Algorithm 5} &  \text{\rm View update algorithm based on
Materialized view}\\\hline \text{\rm Input}:& \text{\rm A definite
deductive database}~ DDB=IDB\cup EDB\cup IC\\
\text{\rm Output:}&\text{\rm A new database}~IDB\cup EDB'\cup IC\\
\text{\rm begin}&\\
~~1.&\text{\rm Let}~ V :=\{ c\in IC~|~IDB\cup IC~\text{\rm
inconsistent
with}~\mathcal{A}~\text{\rm wrt}~c~\}\\
&\text{\rm While}~(V\neq \emptyset)\\
~~2.&\hspace{-0.7cm}\text{\rm For every successful branch $i$:construct}~\Delta_{i}=\{D~|~D\in EDB \}\\
&\text{\rm and D is used as an input clause in branch $i$}.\\
&\text{\rm Construct a branch i of an update tableau based on view}\\
&\text{\rm for}~ IDB\cup EDB~\text{\rm and delete request}~\neg A.\\
&\text{\rm  Produce}~IDB\cup EDB \backslash HS(i)~\text{\rm as a result}\\
~~3.&\text{\rm For every unsuccessful branch $j$:construct}~\Delta_{j}=\{D~|~D\in EDB \}\\
&\text{\rm and D is used as an input clause in branch $j$}.\\
&\text{\rm Construct a branch j of an update magic set rule
based on view}\\
&\text{\rm for}~ IDB\cup EDB~\text{\rm and insert request}~A.\\
&\text{\rm  Produce}~IDB\cup EDB \backslash HS(j)~\text{\rm as a
result}\\
&\text{\rm Let}~ V :=\{ c\in IC~|~IDB\cup IC~\text{\rm inconsistent
with}~\mathcal{A}~\text{\rm wrt}~c~\}\\
&\hspace{-0.7cm}\text{\rm return}\\
~~5.&\text{\rm Produce}~DDB~\text{\rm as the result.}\\
\text{\rm end.}&\\\hline
\end{array}$$

$$\begin{array}{cc}\hline
\text{\bf Algorithm 6} &  \text{\rm View update algorithm based on
Materialized view}\\\hline \text{\rm Input}:& \text{\rm A
stratifiable deductive database}~ DDB=IDB\cup EDB\cup IC\\
\text{\rm Output:}&\text{\rm A new database}~IDB\cup EDB'\cup IC\\
\text{\rm begin}&\\
~~1.&\text{\rm Let}~ V :=\{ c\in IC~|~IDB\cup IC~\text{\rm
inconsistent
with}~\mathcal{A}~\text{\rm wrt}~c~\}\\
&\text{\rm While}~(V\neq \emptyset)\\
~~2.&\hspace{-0.7cm}\text{\rm For every successful branch $i$:construct}~\Delta_{i}=\{D~|~D\in EDB \}\\
&\text{\rm and D is used as an input clause in branch $i$}.\\
&\text{\rm Construct a branch i of an update tableau satisfying
based on view}\\
&\text{\rm for}~ IDB\cup EDB~\text{\rm and delete request}~ A.\\
&\text{\rm  Produce}~IDB\cup EDB \backslash HS(i)~\text{\rm as a result}\\
~~3.&\text{\rm For every unsuccessful branch $j$:construct}~\Delta_{j}=\{D~|~D\in EDB \}\\
&\text{\rm and D is used as an input clause in branch $j$}.\\
&\text{\rm Construct a branch j of an update magic set rule
based on view}\\
&\text{\rm for}~ IDB\cup EDB~\text{\rm and insert request}~A.\\
&\text{\rm  Produce}~IDB\cup EDB \backslash HS(j)~\text{\rm as a
result}\\
&\text{\rm Let}~ V :=\{ c\in IC~|~IDB\cup IC~\text{\rm inconsistent
with}~\mathcal{A}~\text{\rm wrt}~c~\}\\
&\hspace{-0.7cm}\text{\rm return}\\
~~5.&\text{\rm Produce}~DDB~\text{\rm as the result.}\\
\text{\rm end.}&\\\hline
\end{array}$$

Given a database and a view update request, we first transform the
database wrt its Least Herbrand Model (computation of the Least
Herbrand Model can be done as a offline preprocessing step. Note
that it serves as materialized view for efficient query answering).
Then the hyper tableaux calculus (magic set rule) is used to compute
models of this transformed program. Each model represents a rational
way of accomplishing the given view update request. This is
formalized in Algorithms 5 and 6.

This approach for view update may not satisfy (KB*7.1) in general.
But, as shown in the sequel, conformation to (KB*6.3) is guaranteed
and thus this approach results in rational update.

\begin{lemma} Let $T$ be an update tableau based on a materialized view for
$IDB\cup  EDB$ and delete request $\neg A$ ($A$), Similarly, let $M$
be an update magic set rule. Let $S$ be the set of all $EDB$-closed
locally minimal abductive explanations for $A$ wrt $IDB$. Let $S'$
be the set of all $EDB$-cuts of $T$ and  $EDB$-seeds of $M$. Then,
the following hold:
\begin{enumerate}
\item[$\bullet$] $S\subseteq S'$.
\item[$\bullet$] $\forall \Delta'\in S':\exists \Delta\in S ~ s.t.~ \Delta\subseteq \Delta'$.
\item[$\bullet$] $\forall \Delta'\in S':\Delta'\subseteq \bigcup S$.
\end{enumerate}
\end{lemma}

\begin{lemma} Let $S$ and $S'$ be sets of sets s.t. $S\in  S'$ and for every
member $X$ of $S'\backslash S$: $X$ contains a member of $S$ and $X$
is contained in $\bigcup S$. Then, a set $H$ is a hitting set for
$S$ iff it is a hitting set for $S'$.
\end{lemma}

\begin{lemma} Let $T$ and $M$ as in Lemma 5. Then $HS(b)$ is a rational hitting set
for $A$, for every open finished branch $b$ in $T$ (close finished
branch $b$ in $M$).
\end{lemma}

\begin{theorem} Algorithms 5 and 6 are rational, in the sense that they satisfy all
the rationality postulates (KB*1) to (KB*6) and (KB*7.3).
\end{theorem}

\section{Related Works}

We begin by recalling previous work on view deletion. Chandrabose
\cite{Arav1,Arav} and Delhibabu \cite{Del,Del1}, defines a
contraction and revision operator in view deletion with respect to a
set of formulae or sentences using Hansson's \cite{Hans2} belief
change. Similar to our approach, he focused on set of formulae or
sentences in knowledge base revision for view update wrt. insertion
and deletion and formulae are considered at the same level.
Chandrabose proposed different ways to change knowledge base via
only database deletion, devising particular postulate which is shown
to be necessary and sufficient for such an update process.

Our Horn knowledge base consists of two parts, immutable part and
updatable part, but our focus is on minimal change computations.
There is more related works on that topic. Eiter \cite{Eit},
Langlois\cite{Lang}, and Delgrande \cite{Delg} are focusing on Horn
revision with different perspectives like prime implication, logical
closure and belief level. Segerberg \cite{Seg} defined a new
modeling technique for belief revision in terms of irrevocability on
prioritized revision. Hansson \cite{Hans2}, constructed five types
of non-prioritized belief revision. Makinson \cite{Mak} developed
dialogue form of revision AGM. Papini\cite{Pap} defined a new
version of knowledge base revision. In this paper, we considered the
immutable part as a Horn clause and the updatable part as an atom
(literal).

Hansson's\cite{Hans2} kernel change is related to abductive method.
Aliseda's \cite{Alis} book on abductive reasoning is one of the
motivation keys. Christiansen's \cite{Chris} work on dynamics of
abductive logic grammars exactly fits our minimal change (insertion
and deletion). Wrobel's \cite{Wrob} definition of  first order
theory revision was helpful to frame our algorithm.

On other hand, we are dealing with view update problem. Keller's
\cite{Kell} thesis is motivation the view update problem. There is a
lot of papers on the view update problem (for example, the recent
survey paper on view updating by Chen and Liao \cite{Chen} and the
survey paper on view updating algorithms by Mayol and Teniente
\cite{Mayol}  and current survey paper on view selection
(\cite{Potter,Amir,Baum}). More similar to our work is the paper
presented by Bessant et al. \cite{Bess},  which introduces a local
search-based heuristic technique that empirically proves to be often
viable, even in the context of very large propositional
applications. Laurent et al.\cite{Lau}, considers updates in a
deductive database in which every insertion or deletion of a fact
can be performed in a deterministic way.

Furthermore, and at a first sight more related to our work, some
work has been done on ontology systems and description logics (Qi
and Yang \cite{Qi},Kogalovsky \cite{Kog} and Zang \cite{Zhang}). In
Fuzzy related work (\cite{Xu,Lin1,Lin2,Wang,Papa}) also in the
current attenuation of database people.

The significance of our work can be summarized in the following:
\begin{description}
  \item[-]  We have defined a new kind of kernel operator on knowledge bases and obtained
 an axiomatic characterization for it. This operator of change is based on $\alpha$ consistent-remainder set.
  Thus, we have presented a way to construct a kernel operator without the need to make use of the generalized Levi's identity nor
  of a previously defined revision operator.
  \item[-] We have defined a new way of insertion and deletion of an atom(literals) as per
  norm of principle of minimal change.
  \item[-] We have proposed a new generalized revision algorithm for
  knowledge base dynamics, interesting connections with kernel change and
  abduction procedure.
   \item[-] We have designed a new view update algorithm for stratifiable DDB,
 using an axiomatic method based on Hyper tableaux and magic sets.

\end{description}


\section{Conclusion and remarks}

The main contribution of this research is to provide a link between
theory of belief dynamics and concrete applications such as view
updates in databases. We argued for generalization of belief
dynamics theory in two respects: to handle certain part of knowledge
as immutable; and dropping the requirement that belief state be
deductively closed. The intended generalization was achieved by
introducing the concept of knowledge base dynamics and generalized
revision for the same. Further, we also studied the relationship
between knowledge base dynamics and abduction resulting in a
generalized algorithm for revision based on abductive procedures. We
also successfully demonstrated how knowledge base dynamics can
provide an axiomatic characterization for updating an atom(literals)
to a stratifiable (definite) deductive database.

In bridging the gap between belief dynamics and view updates, we
have observed that a balance has to be achieved between
computational efficiency and rationality. While rationally
attractive notions of generalized revision prove to be
computationally inefficient, the rationality behind efficient
algorithms based on incomplete trees is not clear at all. From the
belief dynamics point of view, we may have to sacrifice some
postulates, vacuity for example, to gain computational efficiency.
Further weakening of relevance has to be explored, to provide
declarative semantics for algorithms based on incomplete trees.

On the other hand, from the database side, we should explore various
ways of optimizing the algorithms that would comply with the
proposed declarative semantics. We believe that partial deduction
and loop detection techniques, will play an important role in
optimizing algorithms of the previous section. Note that, loop
detection could be carried out during partial deduction, and
complete SLD-trees can be effectively constructed wrt a partial
deduction (with loop check) of a database, rather than wrt database
itself. Moreover, we would anyway need a partial deduction for
optimization of query evaluation.

We have presented two variants of an algorithm for update a view
atom from a definite database. The key idea of this approach is to
transform the given database into a disjunctive logic program in
such a way that updates can be read off from the models of this
transformed program. One variant based on materialized views is of
polynomial time complexity. Moreover, we have also shown that this
algorithm is rational in the sense that it satisfies the rationality
postulates that are justified from philosophical angle.

In the second variant, where materialized view is used for the
transformation, after generating a hitting set and removing
corresponding $EDB$ atoms, we easily move to the new materialized
view. An obvious way is to recompute the view from scratch using the
new $EDB$ (i.e., compute the Least Herbrand Model of the new updated
database from scratch) but it is certainly interesting to look for
more efficient methods.

Though we have discussed only about view updates, we believe that
knowledge base dynamics can also be applied to other applications
such as view maintenance, diagnosis, and we plan to explore it
further (see works \cite{Caro} and \cite{Bis}). It would also be
interesting to study how results using soft stratification
\cite{Beh} with belief dynamics, especially the relational approach,
could be applied in real world problems. Still, a lot of
developments are possible, for improving existing operators or for
defining new classes of change operators. As immediate extension,
question raises: is there any \emph{real life application for AGM in
25 year theory?} \cite{Ferme}. The revision and update are more
challenging in logical view update problem(database theory), so we
can extend the theory to combine results similar to Konieczny's
\cite{Kon} and Nayak's \cite{Nayak2}.

\section*{Appendix}
\hspace{0.5 cm}\text{Proof of Theorem 1}. Follows from Algorithm 1
and 2. $\blacksquare$

\vspace{0.1cm}

\text{Proof of Theorem 2}. Follows from the result of \cite{Beh}
$\blacksquare$

\vspace{0.1cm}

\text{Proof of Lemma 1}. Follows from the result of \cite{Arav2}

\vspace{0.1cm}

 \text{Proof of Lemma 2 and 5}.
\begin{enumerate}
\item[1.] Consider a $\Delta (\Delta\in\Delta_i \cup \Delta_j)\in S$.
 We need to show that $\Delta$ is generated by algorithm
3 at step 2. It is clear that there exists a $A$-kernel $X$ of
$DDB_G$ s.t. $X \cap EDB = \Delta_j$ and $X \cup EDB = \Delta_i$.
Since $X \vdash A$, there must exist a successful derivation for $A$
using only the elements of $X$ as input clauses and similarly $X
\nvdash A$. Consequently $\Delta   $ must have been constructed at
step 2.
\item[2.] Consider a $\Delta'((\Delta'\in\Delta_i \cup \Delta_j)\in S'$. Let $\Delta'$ be
constructed from a successful(unsuccessful) branch $i$ via
$\Delta_i$($\Delta_j$). Let $X$ be the set of all input clauses used
in the refutation $i$. Clearly $X\vdash A$($X\nvdash A$). Further,
there exists a minimal (wrt set-inclusion) subset $Y$ of $X$ that
derives $A$ (i.e., no proper subset of $Y$ derives $A$). Let $\Delta
= Y \cap EDB$ ($Y \cup EDB$). Since IDB does not(does) have any unit
clauses, $Y$ must contain some EDB facts, and so $\Delta$ is not
empty (empty) and obviously $\Delta\subseteq \Delta'$. But, $Y$ need
not (need) be a $A$-kernel for $IDB_G$ since $Y$ is not ground in
general. But it stands for several $A$-kernels with the same
(different) EDB facts $\Delta$ in them. Thus, from lemma 1, $\Delta$
is a DDB-closed locally minimal abductive explanation for $A$ wrt
$IDB_G$ and is contained in $\Delta'$.
\end{enumerate}

\vspace{0.1cm}

\text{Proof of Lemma 3 and 6}.
\begin{enumerate}
\item[1.] (\textbf{Only if part})~Suppose $H$ is a minimal hitting set for $S$. Since $S
\subseteq S'$ , it follows that $H \subseteq \bigcup S'$ . Further,
$H$ hits every element of $S'$ , which is evident from the fact that
every element of $S'$ contains an element of $S$. Hence $H$ is a
hitting set for $S'$ . By the same arguments, it is not difficult to
see that $H$ is minimal for $S'$ too.\\

(\textbf{If part})~Given that $H$ is a minimal hitting set for $S'$
, we have to show that it is a minimal hitting set for $S$ too.
Assume that there is an element $E \in H$ that is not in $\bigcup
S$. This means that $E$ is selected from some $Y \in S'\backslash
S$. But $Y$ contains an element of $S$, say $X$. Since $X$ is also a
member of $S'$ , one member of $X$ must appear in $H$. This implies
that two elements have been selected from $Y$ and hence $H$ is not
minimal. This is a contradiction and hence $H \subseteq \bigcup S$.
Since $S \subseteq S'$ , it is clear that $H$ hits every element in
$S$, and so $H$ is a hitting set for $S$. It remains to be shown
that $H$ is minimal. Assume the contrary, that a proper subset $H'$
of $H$ is a hitting set for $S$. Then from the proof of the only if
part, it follows that $H'$ is a hitting set for $S'$ too, and
contradicts the fact that $H$ is a minimal hitting set for $S'$ .
Hence, $H$ must be a minimal hitting set for $S$.\\

\item[2.] (\textbf{If part})~Given that $H$ is a hitting set for $S'$ , we have to
show that it is a hitting set for $S$ too. First of all, observe
that $\bigcup S = \bigcup S'$ , and so $H \subseteq \bigcup S$.
Moreover, by definition, for every non-empty member $X$ of $S'$ , $H
\cap X$ is not empty. Since $S \subseteq S'$ , it follows that $H$
is a hitting set for $S$ too.\\

(\textbf{Only if part})~Suppose $H$ is a hitting set for $S$. As
observed above, $H \subseteq \bigcup S'$ . By definition, for every
non-empty member $X\in S$, $X \cap H$ is not empty. Since every
member of $S'$ contains a member of $S$, it is clear that $H$ hits
every member of $S'$ , and hence a hitting set for $S'$ .
$\blacksquare$
\end{enumerate}

\vspace{0.1cm}

\text{Proof of Lemma 4 and 7}. Follows from the lemma 2,3 (minimal
test) and 5,6 (materialized view) of \cite{Beh} $\blacksquare$

\vspace{0.1cm}

 \text{Proof of Theorem 3}. Follows from Lemma 4 and Theorem 1. $\blacksquare$

\vspace{0.1cm}

\text{Proof of Theorem 4}. Follows from Lemma 7 and Theorem 3.
$\blacksquare$


\section*{Acknowledgement}

The author acknowledges the support of RWTH Aachen, where he is
visiting scholar with an Erasmus Mundus External Cooperation Window
India4EU by the European Commission when the paper was written. I
would like to thanks Chandrabose Aravindan and Gerhard Lakemeyer
both my Indian and Germany PhD supervisor, give encourage to write
the paper.

%


\begin{thebibliography}{47}
\bibitem [1] {Alch} Alchourron, C.E., et al.(1985). On the logic of theory change: Partial meet
contraction and revision functions. \emph{Journal of Symbolic Logic
50}, 510 - 530.

\bibitem[2]{Alis} Aliseda, A. (2006). \emph{Abductive Resoning Logic Investigations into
Discovery and Explanation}. Springer book series Vol. 330.

\bibitem[3] {Amir} Amirkhani, H. $\&$ Rahmati. M. (2014). Agreement/disagreement based crowd labeling,
\emph{Applied Intelligence}, Accepted.

\bibitem[4] {Bau} Baumgartner, P., et al. (1997). Semantically Guided Theorem Proving for Diagnosis Applications. \emph{IJCAI} \bf1\rm,
460-465.

\bibitem[5] {Baum} Baumeister, J., et al. (2011). KnowWE: a Semantic Wiki for knowledge engineering,
\emph{Applied Intelligence}, \bf 35\rm (3),  323-344.

\bibitem[6]{Beh} Behrend, A.,$\&$ Manthey,R. (2008). A Transformation-Based
Approach to View Updating in Stratifiable Deductive Databases.\emph{
FoIKS}, 253-271.

\bibitem[7]{Bess} Bessant, B., et al.(1998). Combining Nonmonotonic Reasoning and
Belief Revision: A Practical Approach. \emph{AIMSA}, 115-128.

\bibitem[8] {Bis} Biskup, J.  (2012). Inference-usability confinement by maintaining inference-proof views of an information system.
\emph{IJCSE}. \bf 7 \rm (1), 17-37.

\bibitem[9]{Caro} Caroprese, L., et al.(2012). The View-Update Problem for
Indefinite Databases. \emph{JELIA}.

\bibitem[10]{Arav1} Chandrabose, A.,$\&$ Dung, P.M.(1994). Belief Dynamics, Abduction, and
Database. \emph{JELIA}, 66-85.

\bibitem[11]{Arav} Chandrabose A.(1995), \emph{Dynamics of Belief: Epistmology,
Abduction and Database Update}. Phd Thesis, AIT.

\bibitem[12]{Arav2} Chandrabose. A., $\&$ Baumgartner. P. (1997). A Rational and Efficient Algorithm for View Deletion in Databases.
\emph{ILPS}, 165-179.

\bibitem[13] {Cal} Calvanese, D., et al. (2012). View-based query answering in Description Logics
Semantics and complexity. \emph{J. Comput. Syst. Sci} \bf 78\rm(1),
26-46.

\bibitem[14]{Chen} Chen, H., $\&$ Liao, H. (2010). A Comparative Study of View
Update Problem. \emph{DSDE}, 83-89.

\bibitem[15]{Chris} Christiansen, H., $\&$ Dahl,V. (2009). Abductive Logic
Grammars. \emph{WoLLIC}, 170-181.

\bibitem[16] {Cong} Cong, G., et al. (2012). On the Complexity of View Update Analysis
and Its Application to Annotation Propagation. \emph{IEEE Trans.
Knowl. Data Eng.}\bf 24\rm(3), 506-519.

\bibitem[17]{Delg} Delgrande, J.P, $\&$ Peppas, P. (2011). Revising Horn
Theories. \emph{IJCAI}, 839-844.

\bibitem[18] {Del} Delhibabu, R., $\&$ Lakemeyer, G. (2013). A Rational and Efficient Algorithm for View Revision in
Databases. \emph{Applied Mathematics \& Information Sciences}, \bf
7\rm, pp: 843-856.

\bibitem[19] {Del1} Delhibabu, R. (2014). An Abductive Framework for Knowledge Base Dynamics.
\emph{Applied Mathematics \& Information Sciences} (accepted).


\bibitem[20]{Eit} Eiter, T., $\&$ Makino,K. (2007). On computing all abductive
explanations from a propositional Horn theory. \emph{J. ACM} \bf 54
\rm(5).

\bibitem[21]{Fal} Falappa, M.A., et al.(2012). Prioritized and Non-prioritized
Multiple Change on Belief Bases. \emph{J. Philosophical Logic} \bf
41 \rm(1), 77-113.


\bibitem[22]{Ferme} Fermé, E.L., $\&$ Hansson, S.O. (2011). AGM 25 Years -
Twenty-Five Years of Research in Belief Change. \emph{J.
Philosophical Logic} \bf 40 \rm(2),295-331.

\bibitem[23] {Hans1} Hansson, S.O. (1991). Belief contraction without recovery. \emph{Studia Logica} \bf 50\rm(2),
251 – 260.

\bibitem[24]{Hans2} Hansson, S.O. (1997).\emph{A Textbook of Belief Dynamics}. Kluwer
Academic Publishers, Dordrecht.

\bibitem[25] {Herz} Herzig, A. $\&$ Rifi,O. (1999). Propositional Belief Base
Update and Minimal Change. \emph{Artif. Intell.} \bf 115\rm(1),
107-138.

\bibitem[26] {Jack} Jackson, E. K. $\&$ Schulte, W. (2008). Model Generation for Horn Logic with Stratified
Negation. FORTE.

\bibitem[27] {Kell} Keller, A. (1985).\emph{Updating Relational Databases Through
Views}. Phd Thesis.

\bibitem[28] {Kog} Kogalovsky, M.R. (2012). Ontology-based data access systems. \emph{Programming and Computer Software}
 \bf 38\rm(4), 167-182.

\bibitem[29] {Kon} Konieczny, S. (2011). Dynamics of Beliefs. \emph{SUM},
61-74.

\bibitem[30] {Lak} Lakemeyer, G. (1995). A Logical Account of Relevance.
\emph{IJCAI} (1), 853-861.


\bibitem[31] {Lang}  Langlois, M., et al. (2008). Horn Complements: Towards
Horn-to-Horn Belief Revision. \emph{AAAI}, 466-471.

\bibitem[32] {Lau} Laurent, D., et al. (1998). Updating Intensional Predicates in
Deductive Databases. \emph{Data Knowl. Eng.} \bf 26\rm(1), 37-70.

\bibitem[33] {Lib1} Liberatore, P. (1997).  The Complexity of Belief Update (Extended in 2003). \emph{IJCAI}(1),
68-73.

\bibitem[34] {Lib2} Liberatore, P., $\&$ Schaerf, M. (2004). The Compactness of Belief Revision and Update Operators.
\emph{Fundam. Inform.} \bf 62\rm(3-4), 377-393.

\bibitem[35] {Lin1} Lin, J., et al. (2013). Storing and querying fuzzy XML data in relational databases,
\emph{Applied Intelligence}, \bf 39\rm (2), 386-396.

\bibitem[36] {Lin2} Lin, J. $\&$ Ma. M. (2013). Formal transformation from fuzzy object-oriented databases to fuzzy XML,
\emph{Applied Intelligence}, \bf 39\rm (3), 630-641.

\bibitem[37] {Lobo1} Lobo, J., Minker, J., $\&$ Rajasekar, A. (1992). Foundations of Disjunctive Logic Programming.
\emph{MIT Press, Cambridge}.

\bibitem[38] {Lobo} Lobo, J., $\&$ Trajcevski, G. (1997).  Minimal and Consistent
Evolution of Knowledge Bases, \emph{Journal of Applied Non-Classical
Logics} \bf 7\rm(1).

\bibitem[39] {Lu}  Lu, W. (1999).  View Updates in Disjunctive Deductive Databases
Based on SLD-Resolution. \emph{KRDB}, 31-35.

\bibitem[40] {Mak} Makinson, D. (1997). Screened Revision, \emph{Theoria} \bf 63\rm, 14-23.

\bibitem[41] {Mayol} Mayol, E., $\&$ Teniente, E. (1999). A Survey of Current Methods
for Integrity Constraint Maintenance and View Updating. \emph{ER
(Workshops)}, 62-73.


\bibitem[42] {Meyden} Meyden, R. (1998). Logical Approaches to Incomplete
Information: A Survey. \emph{Logics for Databases and Information
Systems}, 307-356.

\bibitem[43] {Mota} Mota-Herranz, L., et al. (2000). Transaction Trees for Knowledge
Revision, \emph{FQAS}, 182-191.

\bibitem[44] {Nayak1} Nayak, A., et al. (2006).  Forgetting and Knowledge
Update. \emph{Australian Conference on Artificial Intelligence},
131-140.

\bibitem[45] {Nayak2} Nayak, A. (2011).  Is Revision a Special Kind of
Update? \emph{Australasian Conference on Artificial Intelligence},
432-441.

\bibitem[46] {Nebel} Nebel, B. (1998). How Hard is it to Revise a Belief Base?
\emph{Handbook of Defeasible Reasoning and Uncertainty Management
Systems}, 77-145.

\bibitem[47] {Papa} Papadakis, N., et al. (2012). The ramification problem in temporal databases: a solution implemented in SQL,
\emph{Applied Intelligence}, \bf 36\rm (4), 749-767.

\bibitem[48] {Pap} Papini, O.(2000). Knowledge-base revision. \emph{The Knowledge
Engineering Review} \bf 15\rm(4), 339 - 370.

\bibitem[49] {Potter} Potter, s. (2012). Critical reasoning: AI for emergency response,
\emph{Applied Intelligence}, \bf 37\rm (3),  337-356.

\bibitem[50] {Qi} Qi, G., $\&$ Yang, F. (2008). A Survey of Revision Approaches
in Description Logics. \emph{Description Logics}.

\bibitem[51] {Sak} Sakama, C., $\&$ Inoue, K. (2003). An abductive framework for computing
knowledge base updates. \emph{TPLP} \bf3\rm(6), 671-713.

\bibitem[52] {Schul} Schulte, O. (1999).  Minimal Belief Change and
Pareto-Optimality. \emph{ Australian Joint Conference on Artificial
Intelligence}, 144-155.

\bibitem[53] {Seg} Segerberg, K. (1998).  Irrevocable Belief Revision in Dynamic
Doxastic Logic. \emph{Notre Dame Journal of Formal Logic} \bf
39\rm(3), 287-306.

\bibitem[54] {Ten} Teniente, E., $\&$ Urp\'{i}, T. (2003). On the abductive or deductive nature
of database schema validation and update processing problems.
\emph{TPLP} \bf 3 \rm (3), 287-327.

\bibitem[55] {Wang} Wang, J., et al. (2012). On the combination of logical and probabilistic models for information analysis,
\emph{Applied Intelligence}, \bf 36\rm (2), 472-497.

\bibitem[56] {Wrob} Wrobel, S. (1995). First order Theory Refinement. \emph{IOS Frontier
in AI and Application Series}.

\bibitem[57] {Xu} Xu, C., et al. (2012). Efficient fuzzy ranking queries in uncertain
databases, \emph{Applied Intelligence}, \bf 37\rm (1), 47-59.

\bibitem[58] {Zhang} Zhang, F, Z. $\&$ Ma. M. (2014). Representing and Reasoning About XML with Ontologies,
\emph{Applied Intelligence}, \bf 40\rm (1), 74-106.

\end{thebibliography}
\end{document}